\documentclass[submission,copyright,creativecommons]{eptcs}
\providecommand{\event}{QPL 2024} 

\usepackage{iftex}
\usepackage{amsmath,amssymb}
\usepackage{braket}

\usepackage{caption}

\usepackage{colortbl}

\usepackage{graphicx}

\usepackage{stmaryrd}

\newcommand{\semantics}[1]{\llbracket #1 \rrbracket}
\newcommand{\lang}[1]{\ensuremath{\textit{#1}}}

\newcommand{\ot}{\otimes}   
\newcommand{\ep}{\varepsilon}

\newcommand{\FHilb}{\mathbf{FHilb}}
\newcommand{\CPM}[1]{\mathbf{CPM}(#1)}
\newcommand{\CPMC}{\CPM{\mathcal{C}}}

\newcommand{\functor}[1]{\mathsf{#1}}
\newcommand{\hilbsem}{\functor{Q}}
\newcommand{\cpmsem}{\functor{S}}
\newcommand{\cpmpure}{\functor{E}}

\allowdisplaybreaks
\newtheorem{dfn}{Definition}

\newtheorem{thm}{Theorem}

\newtheorem {exa}{Example}

\title{Density Matrices for Metaphor Understanding}
\author{Jay Owers
\institute{SEMT, 
University of Bristol\\
Bristol, UK}
\email{jo16726@bristol.ac.uk}
\and
Ekaterina Shutova
\institute{ILLC, FNWI\\ University of Amsterdam\\
Amsterdam,\\ The Netherlands}
\email{e.shutova@uva.nl}
\and 
Martha Lewis
\institute{SEMT,
University of Bristol\\
Bristol, UK}
\institute{Santa Fe Institute\\Santa Fe, NM, USA}
\email{martha.lewis@bristol.ac.uk}
}
\def\titlerunning{Density Matrices for Metaphor}
\def\authorrunning{J. Owers, E. Shutova, M.Lewis}

\begin{document}

\maketitle

\begin{abstract}
    In physics, density matrices are used to represent mixed states, i.e. probabilistic mixtures of pure states. This concept was used to model lexical ambiguity in \cite{Piedeleu2015}. In this paper, we consider metaphor as a type of lexical ambiguity, and examine whether metaphorical meaning can be effectively modelled using mixtures of word senses. We find that modelling metaphor is significantly more difficult than other kinds of lexical ambiguity, but that our best-performing density matrix method outperforms simple baselines as well as some neural language models. 
\end{abstract}

\section{Introduction}
The use of vectors to model word meaning forms the basis of all modern approaches to modelling language. The idea behind this approach is called \emph{distributional semantics} -- using the distributions of words in text to build vectors that encode word meanings. When we have built vectors for words, say we have $\ket{\lang{cat}}$, $\ket{\lang{kitten}}$, and $\ket{\lang{orthodontist}}$, the hope is that words that have similar meanings will be close together in the vector space, and that words that have dissimilar meanings are further apart, where we measure distance between words as the inner product of the normalised vectors, or equivalently as the cosine similarity, i.e., the cosine of the angle between the vectors. So, for example, we should have that:
\[
\braket{cat|kitten} > \braket{cat|orthodontist}
\]

However, representing words as vectors has some clear disadvantages. Firstly, words can have more than one meaning, and a na\"ive approach to building word vectors will try to pack all those meanings into one vector---ending up with a representation somewhere in the middle. Secondly, as well as representing word meanings we also wish to represent phrases, sentences, and paragraphs of text. However, simply representing words as vectors does not give us an obvious way of composing word vectors to produce phrase or sentence vectors. 

There have been a number of approaches to representing ambiguity for word vectors. The task of discriminating different senses of a given word (word sense discrimination or WSD), is often framed as a classification task: given a sentence with a particular word, identify the sense of that word from its context \cite{lesk_automatic_1986,schutze1998,navigli2009word}. An alternative task is to identify the sense of every word in a sentence. WSD can become a complex task, since each word must be disambiguated with respect to each other word in the sentence. One approach is simply to disambiguate word meanings in advance, and learn different vectors for different senses of a word. However, given that any word may have multiple meanings, this approach could generate a large number of possible meaning combinations that need to be disambiguated. Instead, we would like to be able to compose words together, and for a sentence to be disambiguated in the process of composition.  

A key manifestation of ambiguity in language is the use of metaphor. Metaphor is pervasive in speech, with some estimates showing that we use metaphors on average every 3 sentences. As such, language models need to be able to deal with metaphor when it occurs. Many metaphorical uses are what is known as `conventional' metaphor. This means that the metaphor has become entrenched in language, and can therefore be seen as a case of lexical ambiguity. For example, the most basic use of the word \lang{bright} is as applied to colour. However,  we often use this word to mean `intelligent', as in \lang{bright student}. This meaning of the word was created metaphorically from the original meaning of `colourful', but has now become conventionalised in language.

The field of \emph{compositional distributional semantics} \cite{baroni2010,coecke_mathematical_2010,Mitchell2010} looks at systematic ways to compose word vectors together that are guided by our knowledge of grammatical composition. This paper works within the framework proposed by \cite{coecke_mathematical_2010}, which has fundamental links to quantum theory. Within \cite{coecke_mathematical_2010}, the structure of a sentence, together with the meanings of the words in the sentence, can be viewed as a tensor network creating the meaning of the sentence as a whole. This is further  developed and indeed implemented on quantum computers in \cite{lorenz2023qnlp}. 

In the current work, we look at the ability of density matrices to represent the ambiguity represented by conventional metaphor. \cite{Piedeleu2015} show how density matrices and completely positive maps can be integrated with the compositional distributional semantics proposed by \cite{coecke_mathematical_2010}, and these methods are extended in \cite{meyer2020}. The methods developed by \cite{Piedeleu2015} and applied in \cite{meyer2020} are very successful, beating state of the art language models. We investigate whether density matrix methods are as effective in modelling conventional metaphor as they are in standard cases of lexical ambiguity.

In this paper, we firstly (section \ref{sec:dmms}) give an overview of how density matrices have been used in NLP to model lexical entailment and semantics and syntactic ambiguity. In section \ref{sec:DisCo} we review the quantum-inspired approach to NLP originally proposed by \cite{coecke_mathematical_2010} which extends distributional semantics models to include composition, including how density matrices can be used in this compositional framework, and in section \ref{sec:impl} how density matrices can be built automatically from text. We introduce a new dataset to test how well our implementations are able to model metaphor (Section \ref{sec:methods}), and finally, report results on this new dataset (Section \ref{sec:experiments}), finding that metaphor interpretation is a difficult task for all models tested, although some density matrix methods perform above baseline.

\section{Related Work: Density Matrices in Natural Language Processing}
\label{sec:dmms}
Two key uses of density matrices in NLP are 1) to model hyponymy and ambiguity, and 2) to model lexical ambiguity. We summarize research in these areas. 
\paragraph{Density Matrices for Lexical Entailment} Vectors are not well suited for representing hyponymy (is-a) relations between words. However, since density matrices can be ordered using a variant of the L\"owner ordering \cite{vandewetering2017}, they are a candidate for representing these relations. This was investigated in \cite{sadrzadeh2018}, where density matrices were used to model hyponymy between words and phrases. This work models the strength of a hyponymy relation between two words as a function of the KL-divergence between the two matrices of the words. This measure has the nice property that hyponymy relationships between individual words lift to an entailment relationship between the two sentences. For example, given that \lang{clarify} is a hyponym of \lang{explain} and \lang{rule} is a hyponym of \lang{process}, we would expect that the phrase \lang{clarify rule} entails \lang{explain process}. In similar work, \cite{bankova2019} model the relationship of hyponymy as the L\"owner order between two matrices. The L\"owner order states that $A \leqslant B$ iff $0 \leqslant B-A$. \cite{bankova2019} provide a measure of graded hyponymy between two matrices which again lifts to entailment at the sentence level. \cite{lewis2019} proposes a method for building density matrices using information from WordNet together with off-the shelf word vectors such as word2vec or GloVe. Density matrices for words can be composed to form density matrices for phrases and sentences by a variety of operators such as addition, pointwise multiplication, or more complex operators, such as BMult/Phaser \cite{coecke2020,lewis2019,Piedeleu2015} and KMult/Fuzz \cite{coecke2020,lewis2019}. These representations and composition operators work well on a simple entailment task from \cite{kartsaklis2016}.

The use of density matrices to model logical and conversational entailment is developed further in \cite{lewis2020,rodatz2021,shaikh2021}, who develop a notion of negation for density matrices. 
\cite{rodatz2021,shaikh2021} extend these ideas to include the notion of conversational negation \cite{kruszewski2016}. This can be thought of as modelling the acceptability of a sentence such as `That's not a dog, it's a wolf' vs `That's not a dog, it's a rainbow'. If we view negation as purely logical, then these sentences should be viewed as equally acceptable. In contrast, conversational negation provides a set of alternatives. \cite{rodatz2021,shaikh2021} model this by narrowing the subspace spanned by the logical negation to give a set of relevant alternatives. They furthermore provide alternative models of negation that utilize the Moore-Penrose pseudo-inverse of a matrix. In \cite{cuevas2020}, the monotonicity of composition operators with respect to the L\"owner order is investigated. 

Another approach to building density matrices for entailment was developed by \cite{bradley2020}. In this work, the authors model language as a set of sequences $S$ from a given vocabulary, together with a probability distribution over this set. They form the free vector space over the set of sequences $S$, and then form a rank-1 density operator which encodes the probability distribution over sequences that models the language. Reduced density operators may be formed by tracing out over particular subspaces, and they show that in doing so, the hierarchy of subsequences is encoded by the L\"owner order over these density matrices. This means that they can, for example, encode the fact that \lang{black cats} are a subclass of \lang{cats} - something that is not done by the approach outlined in \cite{sadrzadeh2018} or \cite{bankova2019}.

\paragraph{Density Matrices for Ambiguity}
Using density matrices to model ambiguity in natural language is very natural. Density matrices were introduced in quantum physics to encode the notion of a mixed state: the case where the state of a system is not known. In this case, the system is encoded by taking a \emph{mixture} of the possible states that it could be in. In the case of language, if we consider a word on its own, it may have multiple senses, and without context we cannot tell what the meaning of the word is. For example, \lang{table} on its own can mean a piece of furniture, a structure for storing data, the act of presenting a topic, and so on. If we can store multiple different senses of a word in one representation, this can help to represent language.

One of the first uses of density matrices to represent ambiguity in text was \cite{blacoe2013}. This work aims to build representations of words that can encode multiple different kinds of word usage within one representation. For example, consider the word \lang{table}. This word is semantically ambiguous: \lang{table} means something different in `Your dinner is on the table' vs. `The data is in the table'. It is also syntactically ambiguous, since we can use it as a verb: `table a motion'. The standard way of building word vectors forms a superposition of all these senses. \cite{blacoe2013} firstly build a space that takes into account the grammatical role of words, as encoded by dependency relations. They then form density matrices for individual words, based on the grammatical relations that word can participate in. The representations they learn are effective at the word level, but they do not give any methods for composing words together. \cite{correia2020} also use density matrix representations to model syntactic ambiguity. They show to to model alternative syntactic structures within one whole, and show how word representations can be composed.

\cite{Piedeleu2015} provide a thorough and elegant theoretical grounding that describes how to use the categorical compositional methods of \cite{coecke_mathematical_2010} with density matrix representations of words. This will be described fully in subsequent sections. The main idea behind this is that words are represented as probabilistic mixtures of their senses, and that when words are composed to make phrases, the phrase should disambiguate the meaning of the ambiguous word. The amount of ambiguity in a word can be represented as von Neumann entropy. \cite{Piedeleu2015} carry out corpus-based experiments, and show that the von Neumann entropy of word representations reduces in composition, indicating that the words have been disambiguated.

\cite{meyer2020} extend the work of \cite{Piedeleu2015} to provide a means of building density matrices automatically from large scale text corpora. We describe this method in detail later in the paper. \cite{meyer2020} test their density matrix representations on a range of datasets designed to test the ability of compositional models to disambiguate word meanings. Their representations outperform compositional baselines as well as state of the art large neural models.

\section{Categorical Compositional Distributional Semantics}
\label{sec:DisCo}
We work in the framework of categorical compositional distributional semantics \cite{coecke_mathematical_2010}. In brief, words are represented as vectors inhabiting vector spaces that match their grammatical type. Setting a vector space $N$ to be the noun type, and another space $S$ to be the sentence type, we model nouns as vectors, adjectives as linear maps $adj: N\rightarrow N$ and verbs as multilinear maps from copies of $N$ to $S$.

\subsection{Pregroup Grammars}
\label{sec:PregroupGrammars}
In order to describe grammatical structure we use Lambek's pregroup grammars \cite{lambek1999}. A pregroup   $(P, \leq, \cdot, 1, (-)^l, (-)^r)$ is a partially ordered monoid $(P, \leq, \cdot, 1)$ where each element $p\in P$ has a left adjoint $p^l$ and a right adjoint $p^r$, such that the following inequalities hold:
\begin{equation}
  \label{eq:preg}
  p^l\cdot p \leq 1 \leq p\cdot p^l \quad \text{ and } \quad p\cdot p^r \leq 1 \leq p^r \cdot p
\end{equation}
We think of the elements of a pregroup as linguistic types. Concretely, we will use 
 an alphabet $\mathcal{B} = \{n, s\}$. We use the type $s$ to denote a declarative sentence and $n$ to denote a noun. A transitive verb can then be denoted $n^r s n^l$. If a string of words and their types reduces to the type $s$, the sentence is judged grammatical. The sentence \lang{Junpa loves cats} is typed $n~(n^r s n^l)~ n$, and can be reduced to $s$ as follows: 
\[
n~(n^r s n^l)~ n \leq 1\cdot s n^l n \leq 1 \cdot s \cdot 1 \leq s
\]

\subsection{Compositional Distributional Models}
We interpret a pregroup grammar as a compact closed category, a structure shared by the category of finite dimensional Hilbert spaces. We briefly describe here the structure of a compact closed category, but for more details, please see~\cite{coecke_mathematical_2010, preller2011} and the introduction to relevant category theory given in~\cite{coecke2011}.

Distributional vector space models live in the category $\FHilb$ of finite dimensional real Hilbert spaces and linear maps.
$\FHilb$ is compact closed. Each object $V$ is its own dual and the left and right unit and counit morphisms coincide.
Given a fixed basis $\{ \ket{v_i} \}_i$ of $V$, the unit $\eta$ and counit $\epsilon$ are defined as:
\begin{align*}
  \eta : \mathbb{R} \rightarrow V \otimes V :: 1 \mapsto \sum_i \ket{v_i} \otimes \ket{v_i} \qquad
  \epsilon: V\otimes V \rightarrow \mathbb{R} :: \sum_{ij} c_{ij}\ket{v_i} \otimes \ket{v_j} \mapsto  \sum_{i} c_{ii}
\end{align*}

\subsection{Grammatical Reductions in Vector Spaces}
Following \cite{preller2011}, reductions of the pregroup grammar are mapped into the category $\FHilb$ of finite dimensional Hilbert spaces and linear maps using a strong monoidal functor $\hilbsem$ which preserves the compact closed structure:
\[
\hilbsem: \mathbf{Preg} \rightarrow \FHilb
\]
We map noun and sentence types to appropriate finite dimensional vector spaces $\hilbsem(n) = N$ $\hilbsem(s) = S$, and concatenation in $\mathbf{Preg}$ is mapped to the tensor product in $\FHilb$.
Each type reduction $\alpha$ in the pregroup is mapped to a linear map in $\FHilb$. Given a grammatical reduction $\alpha: p_1, p_2, ... p_n \rightarrow s$ and word vectors $\ket{w_i}$ with types $p_i$, a vector representation of the sentence $w_1 w_2 ... w_n$ is given by:
\[
\ket {w_1 w_2 ... w_n} = \hilbsem(\alpha)(\ket{w_1} \otimes \ket{w_2} \otimes ... \otimes \ket{w_n})
\]
We use the inner product to compare meanings of sentences by computing the cosine distance between sentence vectors.
So, if sentence~$s$ has vector representation~$\ket{s}$
and sentence~$s'$ has representation~$\ket{s'}$, their degree of synonymy is given by: 
\begin{equation*}
  \frac{\braket{s | s'}}
       {\sqrt{\braket{s | s}\braket{s' | s'}}}
\end{equation*}

\subsection{Density Matrices in Categorical Compositional Distributional Semantics}
\label{sec:dmds}

Categorical compositional distributional semantics was extended in \cite{Piedeleu2015} to model nouns as density matrices in $N\otimes N$ and adjective and verbs as completely positive maps.

In distributional models of meaning, density matrices have been used in a variety of ways. We consider the meaning of a word $w$ to be given by a collection of unit vectors~$\{\ket{w_i}\}_i$. Each~$\ket{w_i}$ is weighted by~$p_i \in [0,1]$, such that $\sum_i p_i = 1$.
Then the density~operator:
\[
\semantics{w} = \sum_i p_i \ket{w_i}\bra{w_i}
\]
represents the word~$w$. In \cite{Piedeleu2015}, the vectors $\{\ket{w_i}\}_i$ are interpreted as senses of a given word, and we will use this interpretation later in the paper. In \cite{bankova2019,balkir2015,lewis2019}, the vectors $\{\ket{w_i}\}_i$ are interpreted as exemplars of a concept, and in \cite{blacoe2013} the $\{\ket{w_i}\}_i$ are interpreted as instances of use of a word.

\subsection{The CPM Construction}
Applying Selinger's CPM construction~\cite{selinger} to~$\FHilb$ produces a new compact closed category in which the states are positive operators.
This construction has previously been used in a linguistic setting in~\cite{Kartsaklis2015, Piedeleu2015, balkir2015, bankova2019, lewis2019}. Throughout this section~$\mathcal{C}$ denotes an arbitrary $\dag$-compact closed category.
\begin{dfn}[Completely positive morphism~\cite{selinger}]
  A $\mathcal{C}$-morphism~$\varphi: A^* \otimes A \rightarrow B^* \otimes B$ is said to be completely positive if there exists~$C \in \mathsf{Ob}(\mathcal{C})$
  and~$k \in \mathcal{C}(C\otimes A, B)$, such that~$\varphi$ can be written in the form:
\[
 (k_* \otimes k) \circ (1_{A^*} \otimes \eta_C \otimes 1_A)
\] 
\end{dfn}
Identity morphisms are completely positive, and completely positive morphisms are closed under composition in~$\mathcal{C}$, leading to the following:
\begin{dfn}[$\CPMC$ \cite{selinger}]
  If~$\mathcal{C}$ is a $\dag$-compact closed category then~$\CPMC$ is a category with the same objects as~$\mathcal{C}$ and its morphisms are the completely positive morphisms. 
\end{dfn}
The $\dagger$-compact structure required for interpreting language in our setting lifts to~$\CPM{\mathcal{C}}$:
\begin{thm}[Compact Closure~\cite{selinger}]
  $\CPM{\mathcal{C}}$ is also a $\dagger$-compact closed category.
  There is a functor:
  \begin{align*}
    \cpmpure : \mathcal{C} &\rightarrow \CPM{\mathcal{C}}\\
    k &\mapsto k_* \otimes  k
  \end{align*}
  This functor preserves the $\dagger$-compact closed structure, and is faithful ``up to a global phase''.   
\end{thm}

\subsubsection{Sentence Meaning in the category $\CPM{\FHilb}$}
In the vector space model of distributional models of meaning
the movement from syntax to semantics was achieved via a strong monoidal functor~$\hilbsem : \mathbf{Preg} \rightarrow \FHilb$.
Language can be assigned semantics in~$\CPM{\FHilb}$ in an entirely analogous way via a strong monoidal functor:
\begin{equation*}
  \cpmsem: \mathbf{Preg} \rightarrow \CPM{\FHilb}
\end{equation*}
If $w_1,w_2... w_n$ is  a string of words with corresponding grammatical types~$t_i$ in~$\mathbf{Preg}_\mathcal{B}$.
  and the type reduction is given by~$t_1,...t_n \xrightarrow{r} x$ for some~$x \in \mathsf{Ob}(\mathbf{Preg}_\mathcal{B}$, where $\semantics{w_i}$ is the meaning of word~$w_i$ in~$\CPM{\FHilb}$, i.e. a density matrix $\rho_i$. Then the meaning of~$w_1 w_2 ... w_n$ is given by:
\[
\semantics{w_1 w_2 ... w_n} = \cpmsem(r)(\semantics{w_1} \otimes ... \otimes \semantics{w_n})
\]

We render semantic similarity of representations as the generalised inner product, i.e. $\mathrm{Tr}(\semantics{w_1}^\dagger \semantics{w_2})$, as do \cite{Piedeleu2015}. This notion of semantic similarity is modulated by the extent of the ambiguity of a representation. 
If a representation is maximally ambiguous, that is, $\semantics{w_1} = \mathbb{I}/n$, then the similarity of $\semantics{w_1}$ with itself is only $1/n$. We choose to interpret this as reflecting the fact that the meaning of of $\lang{w}_1$ is undetermined without further context, and hence that the two instances of $\lang{w}_1$ being compared could in fact have different senses. 
On the other hand, if $\semantics{w_1}$ is pure, then self-similarity is 1, as expected. 

We now go on to describe how density matrices can be learnt directly from text corpora, and composed to form sentence representations. We will assess these representations in a setting requiring metaphor interpretation.

We now have all the ingredients to derive sentence meanings in~$\CPM{\FHilb}$.
\begin{exa}\em
We firstly show that the results from $\FHilb$ lift to $\CPM{\FHilb}$.
  Let the noun space~$N$ be a real Hilbert space with basis vectors given by~$\{\ket{n_i}\}_i$, where for some  $i$, $\ket{n_i} = \ket{\lang{shoulders}}$
  Let the sentence space be another space~$S$ with basis $\{\ket{s_i}\}_i$. The verb $\ket{\lang{slouch}}$ is given by:
\[
\ket{\lang{slouch}} = \sum_{pq} C_{pq} \ket{n_p} \ot \ket{s_q}
\]
The density matrix for the noun \lang{shoulders} is in fact a pure state given by:
\[
\semantics{\lang{shoulders}} = \ket{n_i}\bra{n_i}
\]
and similarly, $\semantics{\lang{slouch}}$ in $\CPM{\FHilb}$ is:
\[
\semantics{\lang{slouch}} =  \sum_{pqtu} C_{pq}C_{tu} \ket{n_p}\bra{n_t} \otimes \ket{s_q}\bra{s_u}
\]
The meaning of the composite sentence is simply~$(\ep_N \otimes 1_S)$
applied to~$(\semantics{\lang{shoulders}} \otimes \semantics{\lang{slouch}})$.
This corresponds to:
\begin{align*}
  \semantics{\lang{shoulders slouch}} &= \varphi(\semantics{\lang{shoulders}} \otimes \semantics{\lang{slouch}})\\
&= \sum_{qu}C_{iq}C_{iu}\ket{s_q}\bra{s_u}
\end{align*}
This is a pure state corresponding to the vector $\sum_q C_{iq} \ket{s_q}$.
\end{exa}
We can also deal with mixed states. 
\begin{exa}\em
Let the noun space $N$ be a real Hilbert space with basis vectors given by~$\{\ket{n_i}\}_i$. Consider two senses of the word \lang{shoulder} meaning 1) a part of your body and 2) the edge of a road. Let:
\begin{gather*}
\ket{\lang{shoulder}_\lang{body}} = \sum_i a_i\ket{n_i}, \: \ket{\lang{shoulder}_\lang{road}} = \sum_i b_i\ket{n_i}
\end{gather*}
and with the sentence space $S$, consider the word \lang{slump} with the senses \lang{slouch} and \lang{decline} we define:
\begin{align*}
\ket{\lang{slump}_\lang{slouch}} &= \sum_{pqr} C_{pqr} \ket{n_p} \ot \ket{s_q}\\
\ket{\lang{slump}_\lang{decline}} &= \sum_{pqr} D_{pqr} \ket{n_p} \ot \ket{s_q}
\end{align*}
We set:
\begin{align*}
\semantics{\lang{shoulder}} &= \frac{1}{2}(\ket{\lang{shoulder}_\lang{body}}\bra{\lang{shoulder}_\lang{body}} + \ket{\lang{shoulder}_\lang{road}}\bra{\lang{shoulder}_\lang{road}})\\
\semantics{\lang{slump}} &= \frac{1}{2}(\ket{\lang{slump}_\lang{slouch}}\bra{\lang{slump}_\lang{slouch}} + \ket{\lang{slump}_\lang{decline}}\bra{\lang{slump}_\lang{decline}})
\end{align*}
Then, the meaning of the sentence:
\[
s = \lang{Shoulders slump}
\]
is given by:
\[
\semantics{s} = (\ep_N \otimes 1_S \otimes \ep_N)(\semantics{\lang{shoulders}} \otimes \semantics{\lang{slump}}
\]
\end{exa}

In the example above, we have a two word sentence where each word has two interpretations. There are therefore 4 possible assignments of senses to the words. However, only one assignment of words makes sense in context. The aim is for the correct senses of each word to be picked out in composition.

\section{Methods}
\label{sec:impl}
\subsection{Learning Density Matrices from Text}
\cite{meyer2020} introduce a method for learning density matrices from text called Multi-sense Word2DM. This is an extension of the word2vec skipgram with negative sampling (SGNS) \cite{mikolov2013} to density matrices. word2vec learns vectors for words by running through a large corpus of text and updating word vectors according to the following objective function with regard to model parameters $\theta$:
\begin{equation}\label{negative_sampling}
    J(\theta) = \log \sigma(\braket{v_t|v_c}) + \sum_{k=1}^K  \log \sigma(-\braket{v_t | v_{k}})
\end{equation} 
where $v_t$ is the embedding of target word, $v_c$ is the embedding of the context word, $v_1, v_2, ..., v_K$ are the embeddings of $K$ negative samples, and $\sigma$ is the logistic function. 
Maximising equation \ref{negative_sampling} adjusts the embeddings of words occurring in the same context to be more similar and adjusts the embeddings of words that don't occur together to be less similar.  Multi-sense Word2DM is a modification of SGNS for density matrices. Instead of learning one vector per word, multiple sense embeddings are learnt and then combined together to form a density matrix. 
Each sense of a word has its own $n$-dimensional embedding. A density matrix can be expressed in terms of the sense embeddings as
    \begin{equation}\label{ms_columns}
        A = BB^\dag = \sum_{i=1}^{m} \ket{b_i}\bra{b_i}
    \end{equation}
    where $\ket{b_1}, ..., \ket{b_m}$ are the columns of $B$ corresponding to different senses. 
Each word is also associated with a single vector $v_w$, which represents it as a context word.
    The following objective function is maximised: 
    \begin{equation}\label{ms_word2dm_negative_sampling}
    J(\theta) = \log \sigma(\braket{b_t|c_t}) + \sum_{k=1}^K  \log \sigma(-\braket{b_t | v_{w_k}})
    \end{equation}
    where $\ket{c_t}$ is the sum of context vectors for all words surrounding the target word and $\ket{b_t}$ is the the embedding for the relevant sense of the target word. We select $\ket{b_t}$ by finding the column of $B_t$ most similar to $\ket{c_t}$ (measured by cosine similarity).
Multi-sense Word2DM explicitly models ambiguity by letting the columns of the intermediary matrix represent the different senses of a word. During training the column closest to the context embedding is selected as the relevant sense embedding and only this column is updated.

\subsection{Composition Methods}
\label{sec:comp_methods}
The methods used by \cite{meyer2020} to build density matrices only build matrices that inhabit a single space $W\otimes W$, rather than the larger spaces needed for verbs and adjectives. Because of this, \cite{meyer2020} use a set of composition methods that can be seen as `lifting' a given word representation to the type needed for composition. 
These are based on methods in \cite{lewis2019,coecke2020}, and we will use these in section \ref{sec:experiments}. We view relational words such as adjectives or verbs as maps that takes nouns as arguments.
The composition methods are as follows, using the example of an adjective modifying a noun:
\begin{itemize}
    \item Add: $\semantics{adj} + \semantics{noun}$
    \item Mult: $\semantics{adj} \odot \semantics{noun}$
    \item Fuzz: $\sum_i p_i \, P_i\,  \semantics{noun}\,  P_i$, where $\sum_i p_i \, P_i\,$ is the spectral decomposition of $\semantics{adj}$
    \item Phaser: $\semantics{adj}^{1/2}\semantics{noun}\semantics{adj}^{1/2}$
\end{itemize}
More complex phrases are combined according to their parse. So, a transitive sentence modified with an adjective is composed as $(\text{subj} (\text{verb}(\text{adj obj})))$. Composing e.g. the sentence \lang{Junpa likes stripy cats} would consist of the following steps:

\begin{align}
 \semantics{\lang{stripy cats}} &= f(\semantics{\lang{stripy}}, \semantics{\lang{cats}})\\
 \semantics{\lang{likes stripy cats}} &= f(\semantics{\lang{likes}}, \semantics{\lang{stripy cats}})\\
\semantics{\lang{Junpa likes stripy cats}} &= f(\semantics{\lang{Junpa}}, \semantics{\lang{likes stripy cats}}) 
\end{align}

\noindent where the composer $f$ can be substituted by any of the composition methods listed above.

\section{Implementation }
\label{sec:methods}
\subsection{Datasets and Tasks}
We  build a novel dataset to test disambiguation in a metaphorical context. The basic structure of the dataset is as follows. Given a target sentence that uses a metaphorical word, we minimally alter the target sentence to provide an apt literal paraphrase and an inapt paraphrase of the sentence. We attempt to replace only the single metaphorically used word, although this is not always possible.
\paragraph{Example}
\begin{itemize}
    \item Target sentence: He showered her with presents
    \item Apt paraphrase: He gave her presents
    \item Inapt paraphrase: He sprinkled her with presents
\end{itemize}
The expectation is that the apt paraphrase is semantically closer to the target sentence than the inapt sentence is. So, given representations $\semantics{target}$, $\semantics{apt}$, $\semantics{inapt}$, $\mathrm{Tr}(\semantics{target}^\dag\semantics{apt}) > \mathrm{Tr}(\semantics{target}^\dag\semantics{inapt})$.

We use the metaphorical sentences and their literal paraphrases from \cite{mohammad2016}, and generate inapt paraphrases as follows. Each target sentence includes one metaphorically used verb. We use WordNet \cite{wordnet} to determine the most commonly used sense of the verb. WordNet is a resource listing words, their different senses (called `synsets'), synonyms within each synset,  and hyponym, hypernym and other relations between words. The synsets are listed in order of frequency, so that the first synset is the most commonly used sense of a word. We expect that the most commonly used sense of a word will be a literal sense, and that taking a synonym from this sense will form an inapt paraphrase, as in the example above. We manually inspect each candidate inapt paraphrase, and if the paraphrase is not inapt, we generate a new paraphrase by looking at synonyms of the hypernym of the metaphorically used word. We generated inapt paraphrases for a total of 171 metaphorical sentences.

We then simplified the sentences to consist of just subject-verb (SV), verb-object (VO) or subject-verb-object (SVO) fragments, and subsequently padded these fragments with pronouns and determiners to generate simplified versions of the full sentences. This usually just involved a shortening of the sentence. We test our models on two versions of the dataset. The dataset in the format of SV/VO/SVO fragments we call the \textbf{short-form dataset}, and the dataset with full but simplifed sentences we call the \textbf{long-form dataset}.

\paragraph{Simplification Procedure}
\begin{itemize}
    \item Original sentence: He \textbf{wasted} his inheritance on his insincere friends.
    \item VO format: waste inheritance
    \item Simplified sentence: She wasted her inheritance
\end{itemize}

\begin{figure}[htbp]
    \centering
    \includegraphics[width=\textwidth]{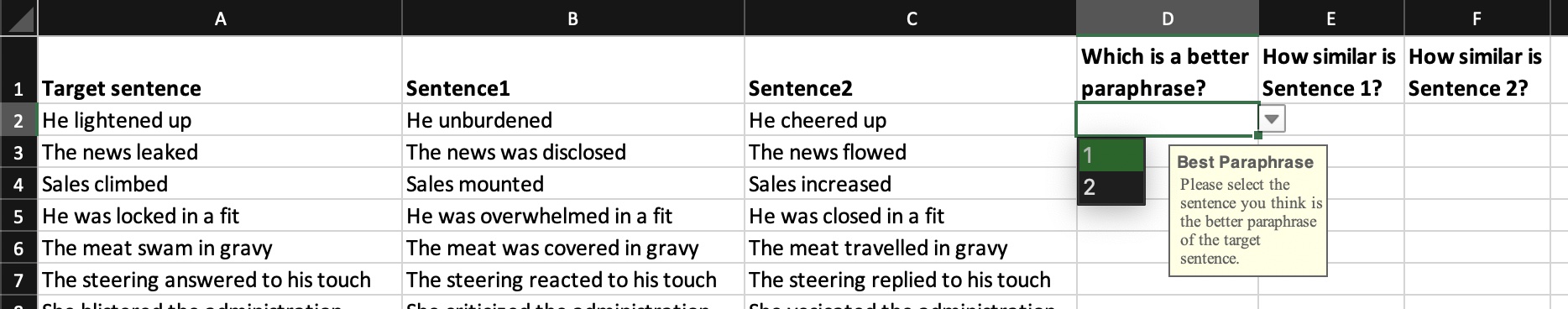}
    \caption{Participants are asked to choose which of paraphrase 1 or 2 is the best.}
    \label{fig:best_paraphrase}
\end{figure}

\begin{figure}
    \centering
    \includegraphics[width=\textwidth]{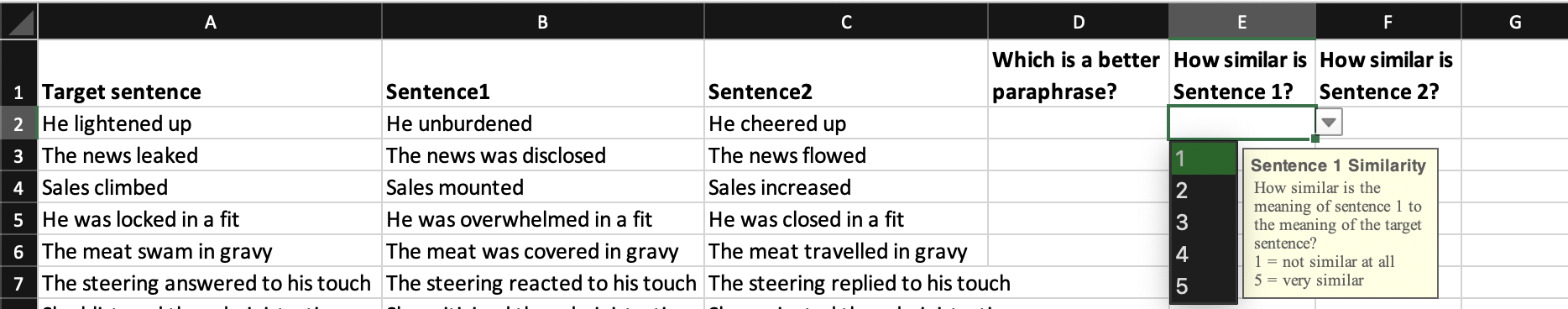}
    \caption{Participants are asked to rate the similarity of each paraphrase to the target sentence}
    \label{fig:similarity}
\end{figure}

\paragraph{Human Annotation}
We solicited annotations from humans as follows. We gave them triples of target sentence, apt paraphrase, and inapt paraphrase. The apt and inapt paraphrases were presented in a random order: sometimes the apt was presented first, and sometimes the inapt. Participants were asked to state which was the best paraphrase of the target sentence, and then to rate the similarity of each paraphrase to the target sentence. All participants were asked to annotate all 171 triples, resulting in 10 annotations per triple. Triples were presented in a random order for each participant. Mean inter-annotator agreement, calculated using Spearman's rho over all pairs of annotators, was 0.589, which is a reasonable level. Examples of the annotation task are shown in figures \ref{fig:best_paraphrase} and \ref{fig:similarity}

\paragraph{Model Task}
We evaluate a range of density matrix models, neural models, and baselines  on this disambiguation task. The evaluation runs as follows. For each triple of sentences, we generate vectors or density matrices for each sentence. We then compute the similarity of the sentence representations using either cosine similarity (for vectors) or the generalized inner product (for density matrices. We assess the extent to which models agree with the mean of the human judgements using Spearman's rho.

\subsection{Models}
\label{sec:models}
We use the density matrices computed in \cite{meyer2020}. We test 4 variants of \textbf{Multi-sense Word2DM} (ms-Word2DM). One parameter varies the number of senses computed for each matrix, either 5 or 10, and one varies the means by which the closest sense is chosen for updating, choosing the closest via cosine similarity (c) or by Euclidean distance (d).\cite{meyer2020} also propose two other key ways of generating density matrices. One, called \textbf{Context2DM}, extends the methods proposed in \cite{Piedeleu2015,schutze1998}. Firstly, a set of word vectors is trained with the gensim implementation\footnote{https://radimrehurek.com/gensim/models/word2vec} of Word2Vec on the combined ukWaC+Wackypedia corpus. For each target word, the set of words whose context they appear in is collected. The vectors of these context words are clustered using hierarchical agglomerative clustering to between 2 and 10 clusters. The centroids of each of these clusters are taken to form the sense vectors for the target word, and subsequently these sense vectors are combined into a density matrix.

The second alternative method, \textbf{Bert2DM} uses BERT \cite{devlin2019} to generate sense vectors. BERT is a Transformer-based architecture \cite{vaswani2017} which produces contextual embeddings, meaning that the vector produced for each word is dependent on the context of the sentence. A small corpus is fed through BERT. Vectors for each word are extracted, and dimensionality reduction applied (either PCA or SVD). Vectors for each word are then combined to form density matrices.

We compare performance with two neural sentence encoders, \textbf{SBERT} \cite{reimers2019} and \textbf{InferSent} \cite{conneau2017}. Both of these models are explicitly trained to predict when one sentence can be inferred from another. Since synonymy is a simple kind of entailment relation, these sentence encoders should perform well.

We compare with two baseline models, \textbf{word2vec} \cite{mikolov2013} and \textbf{GloVe} \cite{pennington2014glove}. These are two methods for producing static word vectors with the key property that semantically similar words should be close together in the vector space.

\section{Experiments and Results}
\label{sec:experiments}
For each model described in the previous section, we generate sentence embeddings for triples in the dataset. For the density matrix methods, we apply the composition methods outlined in \ref{sec:comp_methods}, that is, Add, Mult, Fuzz, and Phaser. For neural methods, we simple take the sentence embeddings produced by the models. For the static word vector baselines, we apply Add and Mult. We also use a baseline `composition' method of simply taking the verb as sentence representation (Verb only)

We compute $sim(\semantics{target}, \semantics{apt})$ and $sim(\semantics{target}, \semantics{inapt})$, where $sim$ is cosine similarity for vectors and generalized inner product for density matrices. We compare the similarity scores generated by the models to the similarity scores generated by humans and compute the correlation between these scores using Spearman's rho. Spearman's rho ranges between -1 and 1, with 1 indicating that scores are perfectly correlated and -1 indicating that they are perfectly anti-correlated.

For each model tested, there were a number of occurrences of words in the dataset that were not in the lexicon of the trained model. Where this occurred, the sentence pairs containing these words were removed from the tests for the given model. The error occurred more frequently with the long-form dataset due to words not being in their lemmatised form. For Context2DM and BERT2DM with the long-form dataset, the results were deemed unreliable and not included due to having only 8 usable sentence pairs. Table \ref{tab:errors} in the Appendix shows the number of sentence pairs used in the tests, out of the total of 342 sentence pairs, as a result of these errors being removed.

\subsection{Results}
Results are shown in tables \ref{tab:vect}, \ref{tab:dm1} and \ref{tab:dm2}. Table \ref{tab:vect} shows firstly that the neural sentence encoders (left hand table) were not able to correctly interpret the metaphorically used word in context, with correlation close to 0, i.e. indicating chance level. Secondly, on the right hand side, we see that in the compositional settings (Add and Mult), static word vectors also perform poorly, although very slightly better than the sentence encoders. For the Verb-only setting, we see a stronger negative correlation, i.e., similarity ratings are more consistently in the wrong direction. This is expected, as the verb alone does not provide the context necessary for disambiguation of the metaphor, and we would expect that the most common sense of a word would dominate the learnt representations. 
\begin{table}[hbtp]
\begin{center}
    \begin{tabular}{rl|r}
    \multicolumn{1}{l}{SBERT} & short & \cellcolor[rgb]{ .984,  .925,  .937}-0.0317 \\
          & long  & \cellcolor[rgb]{ .984,  .957,  .969}-0.0146 \\
    \hline
    \multicolumn{1}{l}{Infersent1} & short & \cellcolor[rgb]{ .945,  .957,  .984}0.0085 \\
          & long  & \cellcolor[rgb]{ .984,  .98,  .992}-0.0027 \\
    \hline
    \multicolumn{1}{l}{Infersent2} & short & \cellcolor[rgb]{ .929,  .949,  .98}0.0113 \\
          & long  & \cellcolor[rgb]{ .984,  .906,  .918}-0.0411 \\
          
    \end{tabular}%
    \hfill
    \begin{tabular}{rl|rrr}
          &       & \multicolumn{1}{l}{Verb} & \multicolumn{1}{l}{Add} & \multicolumn{1}{l}{Mult} \\
\hline    \multicolumn{1}{l}{Word2Vec} & short & \cellcolor[rgb]{ .973,  .463,  .471}-0.2719 & \cellcolor[rgb]{ .984,  .882,  .894}-0.0536 & \cellcolor[rgb]{ .98,  .776,  .788}-0.1092 \\
          & long  &       & \cellcolor[rgb]{ .984,  .984,  .996}-0.0010 & \cellcolor[rgb]{ .984,  .863,  .875}-0.0642 \\
\hline    \multicolumn{1}{l}{GloVe} & short & \cellcolor[rgb]{ .976,  .694,  .702}-0.1529 & \cellcolor[rgb]{ .984,  .933,  .945}-0.0272 & \cellcolor[rgb]{ .6,  .714,  .863}0.0738 \\
          & long  &       & \cellcolor[rgb]{ .867,  .902,  .957}0.0236 & \cellcolor[rgb]{ .682,  .773,  .894}0.0579 \\
    \end{tabular}%
\vspace{0.4cm}
    \begin{tabular}{rrrrrrr}
    \rowcolor[rgb]{ .973,  .412,  .42}  & \cellcolor[rgb]{ .976,  .6,  .612} & \cellcolor[rgb]{ .98,  .796,  .804} & \cellcolor[rgb]{ .988,  .988,  1} & \cellcolor[rgb]{ .776,  .839,  .925} & \cellcolor[rgb]{ .565,  .69,  .851} & \cellcolor[rgb]{ .353,  .541,  .776} \\
    -0.3  & -0.2  & -0.1  & 0     & 0.04  & 0.08  & 0.12 \\
    \end{tabular}%

\end{center}
\caption{Spearman's rho for sentence encoders and vector baselines. Note that all sentence encoders performed poorly, while Glove with mult performed better.}
\label{tab:vect}
\end{table}

In Table \ref{tab:dm1}, we see a similar pattern. The verb-only baseline produces similarity ratings that are negatively correlated with those of humans. In general, the BERT2DM and Context2DM models produce slightly stronger results. The highest correlation is produced by Multi-sense Word2DM with 10 senses, using Euclidean distance to choose which sense to update, and using Mult as the composition operator. Overall, correlation is low.
Table \ref{tab:dm2} gives performance of the density matrix models when using Fuzz or Phaser as the composition operator. When applying Fuzz and Phaser, there is a choice of whether to use the verb as the operator or the noun as the operator. The linguistically motivated choice is to use the verb as the operator. However, as seen in table \ref{tab:dm2}, using the noun as the operator produces a better correlation with human judgements. We speculate that when using the verb as operator, the most common sense of the verb dominates the composition. In the case of a metaphorical verb, we require that the noun modifies the meaning of the verb to obtain the correct interpretation.

\begin{table}[hbtp]
\begin{center}
    \begin{tabular}{ll|rrr}
          &       & \multicolumn{1}{l}{Verb} & \multicolumn{1}{l}{Add} & \multicolumn{1}{l}{Mult} \\
    \hline
    ms-Word2DM-c5 & short & \cellcolor[rgb]{ .976,  .588,  .6}-0.2062 & \cellcolor[rgb]{ .984,  .875,  .886}-0.0587 & \cellcolor[rgb]{ .882,  .914,  .965}0.0201 \\
          & long  &       & \cellcolor[rgb]{ .792,  .851,  .933}0.0373 & \cellcolor[rgb]{ .98,  .839,  .851}-0.0764 \\
    \hline
    ms-Word2DM-c10 & short & \cellcolor[rgb]{ .976,  .584,  .592}-0.2094 & \cellcolor[rgb]{ .984,  .878,  .89}-0.0552 & \cellcolor[rgb]{ .984,  .918,  .925}-0.0365 \\
          & long  &       & \cellcolor[rgb]{ .753,  .824,  .918}0.0450 & \cellcolor[rgb]{ .984,  .875,  .886}-0.0577 \\
    \hline
    ms-Word2DM-d5 & short & \cellcolor[rgb]{ .973,  .502,  .51}-0.2519 & \cellcolor[rgb]{ .98,  .796,  .804}-0.0999 & \cellcolor[rgb]{ .984,  .945,  .957}-0.0204 \\
          & long  &       & \cellcolor[rgb]{ .918,  .941,  .976}0.0136 & \cellcolor[rgb]{ .788,  .847,  .929}0.0385 \\
    \hline
    ms-Word2DM-d10 & short & \cellcolor[rgb]{ .976,  .584,  .592}-0.2102 & \cellcolor[rgb]{ .98,  .835,  .847}-0.0784 & \cellcolor[rgb]{ .984,  .871,  .878}-0.0609 \\
          & long  &       & \cellcolor[rgb]{ .761,  .827,  .922}0.0432 & \cellcolor[rgb]{ .427,  .596,  .804}0.1061 \\
    \hline
    Word2DM & short & \cellcolor[rgb]{ .976,  .647,  .655}-0.1772 & \cellcolor[rgb]{ .98,  .78,  .792}-0.1063 & \cellcolor[rgb]{ .984,  .984,  .996}-0.0004 \\
          & long  &       & \cellcolor[rgb]{ .984,  .976,  .988}-0.0057 & \cellcolor[rgb]{ .851,  .894,  .953}0.0263 \\
    \hline
    bert2dm-pca-cls & short & \cellcolor[rgb]{ .925,  .945,  .98}0.0120 & \cellcolor[rgb]{ .592,  .71,  .863}0.0749 & \cellcolor[rgb]{ .851,  .894,  .953}0.0262 \\
    \hline
    bert2dm-svd-cls & short & \cellcolor[rgb]{ .984,  .929,  .941}-0.0296 & \cellcolor[rgb]{ .898,  .925,  .969}0.0172 & \cellcolor[rgb]{ .62,  .729,  .871}0.0698 \\
    \hline
    context2dm & short & \cellcolor[rgb]{ .863,  .902,  .957}0.0240 & \cellcolor[rgb]{ .733,  .812,  .914}0.0483 & \cellcolor[rgb]{ .522,  .659,  .835}0.0885 \\
    \end{tabular}%
    
\vspace{0.4cm}
    \begin{tabular}{rrrrrrr}
    \rowcolor[rgb]{ .973,  .412,  .42}  & \cellcolor[rgb]{ .976,  .6,  .612} & \cellcolor[rgb]{ .98,  .796,  .804} & \cellcolor[rgb]{ .988,  .988,  1} & \cellcolor[rgb]{ .776,  .839,  .925} & \cellcolor[rgb]{ .565,  .69,  .851} & \cellcolor[rgb]{ .353,  .541,  .776} \\
    -0.3  & -0.2  & -0.1  & 0     & 0.04  & 0.08  & 0.12 \\
    \end{tabular}%
    
\end{center}
\caption{Spearman's rho for density matrix models with simple composition. Most models with verb-only composition had negative correlation. ms-Word2DM-d10 with mult-long had the best performance.}
\label{tab:dm1}
\end{table}

All models except for BERT2DM-svd-cls and Context2DM consistently show an increase in rho after composition, supporting the idea that interpretation of metaphor is easier when provided with context. 

\begin{table}
\begin{center}
    \begin{tabular}{l|rrrr}
          & \multicolumn{1}{l}{Fuzz verb} & \multicolumn{1}{l}{Fuzz noun} & \multicolumn{1}{l}{Phaser verb} & \multicolumn{1}{l}{Phaser noun} \\
\hline    ms-Word2DM-c5 & \cellcolor[rgb]{ .976,  .655,  .663}-0.1732 & \cellcolor[rgb]{ .788,  .851,  .933}0.0378 & \cellcolor[rgb]{ .984,  .937,  .949}-0.0251 & \cellcolor[rgb]{ .914,  .933,  .973}0.0148 \\
    ms-Word2DM-c10 & \cellcolor[rgb]{ .976,  .663,  .675}-0.1681 & \cellcolor[rgb]{ .984,  .957,  .969}-0.0146 & \cellcolor[rgb]{ .98,  .804,  .816}-0.0949 & \cellcolor[rgb]{ .984,  .957,  .969}-0.0160 \\
    ms-Word2DM-d5 & \cellcolor[rgb]{ .976,  .643,  .655}-0.1783 & \cellcolor[rgb]{ .686,  .776,  .894}0.0576 & \cellcolor[rgb]{ .98,  .769,  .78}-0.1132 & \cellcolor[rgb]{ .925,  .945,  .98}0.0121 \\
    ms-Word2DM-d10 & \cellcolor[rgb]{ .976,  .604,  .612}-0.1997 & \cellcolor[rgb]{ .776,  .839,  .925}0.0402 & \cellcolor[rgb]{ .976,  .686,  .698}-0.1552 & \cellcolor[rgb]{ .984,  .961,  .973}-0.0125 \\
    Word2DM & \cellcolor[rgb]{ .98,  .784,  .796}-0.1042 & \cellcolor[rgb]{ .984,  .984,  .996}-0.0008 & \cellcolor[rgb]{ .98,  .733,  .745}-0.1311 & \cellcolor[rgb]{ .984,  .98,  .992}-0.0029 \\
    bert2dm-pca-cls & \cellcolor[rgb]{ .875,  .91,  .961}0.0217 & \cellcolor[rgb]{ .843,  .886,  .949}0.0274 & \cellcolor[rgb]{ .812,  .863,  .937}0.0337 & \cellcolor[rgb]{ .89,  .922,  .969}0.0186 \\
    bert2dm-svd-cls & \cellcolor[rgb]{ .984,  .933,  .945}-0.0280 & \cellcolor[rgb]{ .792,  .851,  .933}0.0377 & \cellcolor[rgb]{ .984,  .898,  .91}-0.0451 & \cellcolor[rgb]{ .8,  .855,  .933}0.0358 \\
    context2dm & \cellcolor[rgb]{ .898,  .925,  .969}0.0177 & \cellcolor[rgb]{ .753,  .824,  .918}0.0448 & \cellcolor[rgb]{ .984,  .969,  .98}-0.0088 & \cellcolor[rgb]{ .804,  .859,  .937}0.0350 \\
    \end{tabular}%
    
    \vspace{0.4cm}
    \begin{tabular}{rrrrrrr}
    \rowcolor[rgb]{ .973,  .412,  .42}  & \cellcolor[rgb]{ .976,  .6,  .612} & \cellcolor[rgb]{ .98,  .796,  .804} & \cellcolor[rgb]{ .988,  .988,  1} & \cellcolor[rgb]{ .776,  .839,  .925} & \cellcolor[rgb]{ .565,  .69,  .851} & \cellcolor[rgb]{ .353,  .541,  .776} \\
    -0.3  & -0.2  & -0.1  & 0     & 0.04  & 0.08  & 0.12 \\
    \end{tabular}%
\end{center}
\caption{Spearman's rho for density matrix models with Fuzz and Phaser composition on the short-form dataset. Verb operator models generally had poor performance while noun operator models, especially with Fuzz, performed better.}
\label{tab:dm2}
\end{table}

\subsection{Analysis}
\begin{figure}[hbtp]
    \centering
    \includegraphics[width=0.8\textwidth]{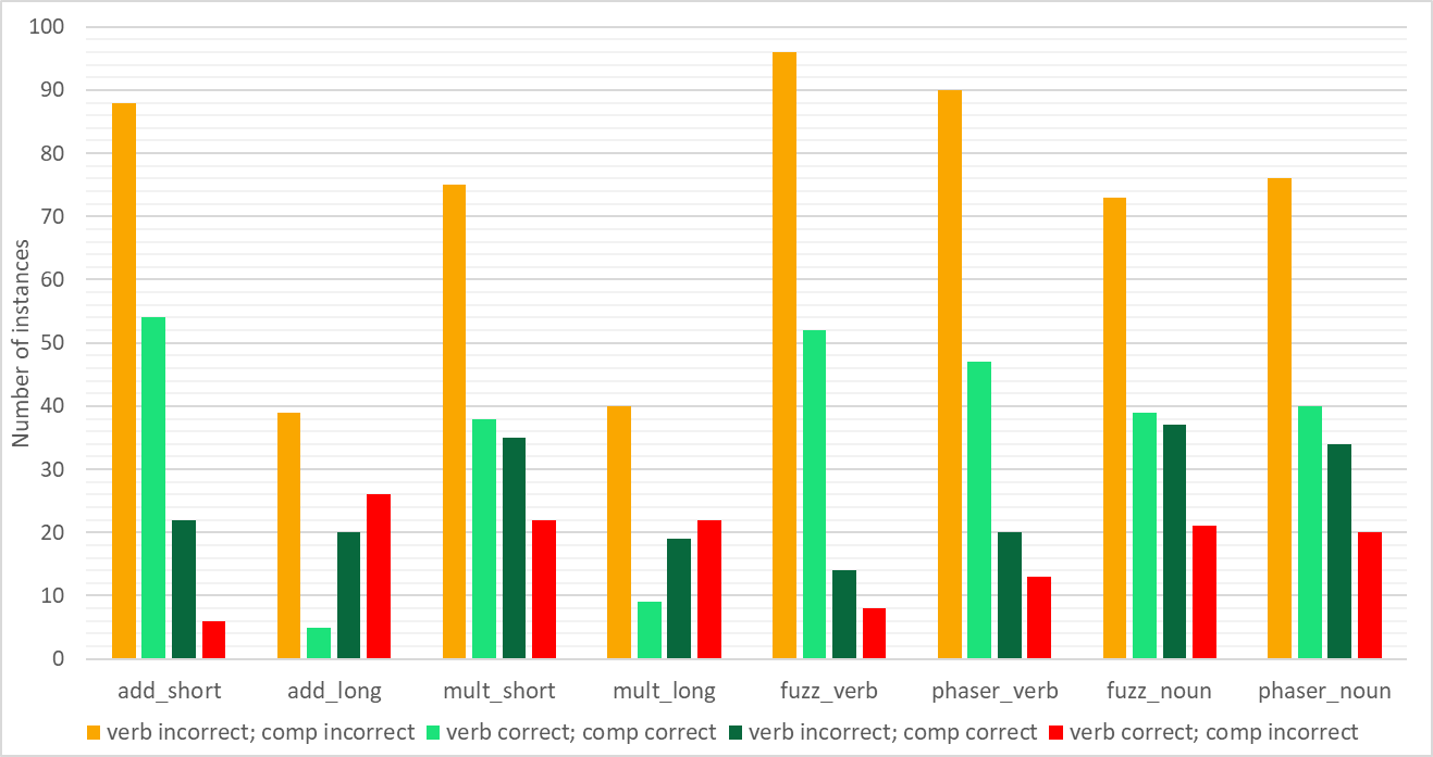}
    \caption{Analysis of which sentences were scored correctly by ms-Word2DM-d10. Note that verb operator models scored a lot of sentences the same as the verb, meaning composition had little effect.}
    \label{fig:sentence analysis}
\end{figure}
To understand more about which sentences were being scored correctly, we examined the responses of ms-word2dm-d10. Each instance of a metaphorical sentence and its two paraphrases was marked correct if the apt paraphrase scored higher on cosine similarity than the inapt paraphrase, and incorrect if the inapt paraphrase scored highest. These results were then compiled for different composition methods and compared with the verb baseline. The number of instances correct and incorrect with verb and composition are shown in figure \ref{fig:sentence analysis}. We see that overall, the majority of similarity judgements are incorrect for verb-only models and for compositional models (yellow bar). We also see that for many models, both the verb-only and the compositional models are correct (light green bar). For Fuzz and Phaser, models where the verb is the operator showed fewer instances where the result changes after composition. This means that in more cases, the result remained either correct or incorrect according to the result of the verb alone. In contrast, the noun operator models had more instances where the verb-only model was incorrect and the composition was correct (dark green bar). This also indicates that the result is being heavily influenced by the operator matrix. Finally, there are a lot of instances where the verb-only model produces the correct answer but this is `undone' by the composition (red bar). For Add and Mult, the long-form sentences showed a greater proportion of the correct sentences using the verb that became incorrect after composition, suggesting that the extra words in the long sentences may be distracting from the relevant information.

To analyse how each composition method affects the modelled ambiguity, we calculated the von Neumann entropy of each of the sentences for each composition method including the verb baseline, as shown in tables \ref{tab:vne1} and \ref{tab:vne2}. The relative difference in entropy between the verb and the sentence embedding indicates the model's change in ambiguity upon composition of the sentence. This is shown by the colour of the cells, red meaning an increase in entropy and blue meaning a decrease. The entropy results are consistent with the results for the similarity task; Phaser resulted in a greater reduction in entropy than Fuzz, and also produced stronger Spearman correlation. BERT2DM-svd-cls mult-short also had a high rho and a strong reduction in entropy.

\begin{table}[hbtp]
\begin{center}
    \begin{tabular}{ll|rrr}
          &       & \multicolumn{1}{l}{Verb} & \multicolumn{1}{l}{Add} & \multicolumn{1}{l}{Mult} \\
    \hline
    ms-Word2DM-c5 & short & \cellcolor[rgb]{ .988,  .988,  1}1.1464 & \cellcolor[rgb]{ .988,  .859,  .871}1.9298 & \cellcolor[rgb]{ .988,  .894,  .902}1.6927 \\
          & long  & \cellcolor[rgb]{ .988,  .988,  1}1.1464 & \cellcolor[rgb]{ .988,  .882,  .894}1.7579 & \cellcolor[rgb]{ .988,  .871,  .882}1.8508 \\
    \hline
    ms-Word2DM-c10 & short & \cellcolor[rgb]{ .988,  .988,  1}1.3230 & \cellcolor[rgb]{ .988,  .863,  .871}2.2118 & \cellcolor[rgb]{ .988,  .929,  .941}1.6821 \\
          & long  & \cellcolor[rgb]{ .988,  .988,  1}1.3230 & \cellcolor[rgb]{ .988,  .89,  .902}1.9797 & \cellcolor[rgb]{ .988,  .89,  .902}1.9587 \\
    \hline
    ms-Word2DM-d5 & short & \cellcolor[rgb]{ .988,  .988,  1}1.0590 & \cellcolor[rgb]{ .984,  .831,  .843}1.9834 & \cellcolor[rgb]{ .988,  .906,  .918}1.4786 \\
          & long  & \cellcolor[rgb]{ .988,  .988,  1}1.0590 & \cellcolor[rgb]{ .988,  .871,  .878}1.7171 & \cellcolor[rgb]{ .988,  .878,  .89}1.6482 \\
    \hline
    ms-Word2DM-d10 & short & \cellcolor[rgb]{ .988,  .988,  1}1.4508 & \cellcolor[rgb]{ .988,  .894,  .902}2.1462 & \cellcolor[rgb]{ .988,  .965,  .976}1.6160 \\
          & long  & \cellcolor[rgb]{ .988,  .988,  1}1.4508 & \cellcolor[rgb]{ .988,  .91,  .922}2.0085 & \cellcolor[rgb]{ .988,  .925,  .937}1.8854 \\
    \hline
    Word2DM & short & \cellcolor[rgb]{ .988,  .988,  1}0.1799 & \cellcolor[rgb]{ .976,  .443,  .455}1.5876 & \cellcolor[rgb]{ .98,  .565,  .573}0.9849 \\
          & long  & \cellcolor[rgb]{ .988,  .988,  1}0.1799 & \cellcolor[rgb]{ .976,  .549,  .557}1.0555 & \cellcolor[rgb]{ .984,  .765,  .776}0.4431 \\
    \hline
    bert2dm-pca-cls & short & \cellcolor[rgb]{ .988,  .988,  1}0.4647 & \cellcolor[rgb]{ .984,  .784,  .796}1.0575 & \cellcolor[rgb]{ .98,  .984,  .996}0.4467 \\
    bert2dm-svd-cls & short & \cellcolor[rgb]{ .988,  .988,  1}0.2346 & \cellcolor[rgb]{ .984,  .804,  .816}0.4943 & \cellcolor[rgb]{ .718,  .8,  .906}0.0339 \\
    context2dm & short & \cellcolor[rgb]{ .988,  .988,  1}0.2877 & \cellcolor[rgb]{ .988,  .875,  .886}0.4546 & \cellcolor[rgb]{ .988,  .973,  .984}0.3097 \\
    \end{tabular}%
    
\vspace{0.4cm}
    \begin{tabular}{rrrrrrr}
    \rowcolor[rgb]{ .353,  .541,  .776}    & \cellcolor[rgb]{ .565,  .69,  .851} & \cellcolor[rgb]{ .773,  .835,  .922} & \cellcolor[rgb]{ .988,  .988,  1} & \cellcolor[rgb]{ .984,  .8,  .812} & \cellcolor[rgb]{ .98,  .604,  .612} & \cellcolor[rgb]{ .973,  .412,  .42} \\
    0.01  & 0.05 & 0.22 & 1  & 2.15  & 4.64  & 10 \\
    \end{tabular}%
\end{center}
\caption{Mean von Neumann entropy produced by density matrix models with simple composition. Colour indicates the relative change in entropy with composition. Bert2dm-svd-cls with mult is the only model to significantly decrease entropy.}
\label{tab:vne1}
\end{table}

\begin{table}
\begin{center}
    \begin{tabular}{l|rrrrr}
          & \multicolumn{1}{l}{Verb} & \multicolumn{1}{l}{Fuzz verb} & \multicolumn{1}{l}{Fuzz noun} & \multicolumn{1}{l}{Phaser verb} & \multicolumn{1}{l}{Phaser noun} \\
    \hline
    ms-Word2DM-c5 & \cellcolor[rgb]{ .988,  .988,  1}1.1464 & \cellcolor[rgb]{ .91,  .933,  .973}0.6781 & \cellcolor[rgb]{ .914,  .933,  .973}0.6731 & \cellcolor[rgb]{ .827,  .875,  .941}0.3661 & \cellcolor[rgb]{ .824,  .871,  .941}0.3550 \\
    ms-Word2DM-c10 & \cellcolor[rgb]{ .988,  .988,  1}1.3230 & \cellcolor[rgb]{ .925,  .945,  .976}0.8769 & \cellcolor[rgb]{ .933,  .949,  .98}0.9141 & \cellcolor[rgb]{ .851,  .894,  .953}0.4971 & \cellcolor[rgb]{ .847,  .89,  .949}0.4860 \\
    ms-Word2DM-d5 & \cellcolor[rgb]{ .988,  .988,  1}1.0590 & \cellcolor[rgb]{ .925,  .945,  .976}0.7257 & \cellcolor[rgb]{ .925,  .941,  .976}0.6873 & \cellcolor[rgb]{ .812,  .863,  .937}0.3046 & \cellcolor[rgb]{ .804,  .855,  .933}0.2798 \\
    ms-Word2DM-d10 & \cellcolor[rgb]{ .988,  .988,  1}1.4508 & \cellcolor[rgb]{ .91,  .929,  .969}0.8580 & \cellcolor[rgb]{ .91,  .933,  .973}0.8412 & \cellcolor[rgb]{ .812,  .863,  .937}0.4034 & \cellcolor[rgb]{ .804,  .859,  .933}0.3918 \\
    Word2DM & \cellcolor[rgb]{ .988,  .988,  1}0.1799 & \cellcolor[rgb]{ .988,  .863,  .875}0.2896 & \cellcolor[rgb]{ .988,  .918,  .929}0.2519 & \cellcolor[rgb]{ .792,  .847,  .929}0.0511 & \cellcolor[rgb]{ .788,  .847,  .929}0.0413 \\
    bert2dm-pca-cls & \cellcolor[rgb]{ .988,  .988,  1}0.4647 & \cellcolor[rgb]{ .922,  .941,  .976}0.3001 & \cellcolor[rgb]{ .933,  .949,  .98}0.3182 & \cellcolor[rgb]{ .804,  .859,  .933}0.1245 & \cellcolor[rgb]{ .808,  .859,  .933}0.1269 \\
    bert2dm-svd-cls & \cellcolor[rgb]{ .988,  .988,  1}0.2346 & \cellcolor[rgb]{ .757,  .824,  .918}0.0444 & \cellcolor[rgb]{ .788,  .847,  .929}0.0546 & \cellcolor[rgb]{ .557,  .682,  .847}0.0103 & \cellcolor[rgb]{ .557,  .682,  .847}0.0103 \\
    context2dm & \cellcolor[rgb]{ .988,  .988,  1}0.2877 & \cellcolor[rgb]{ .824,  .875,  .941}0.0912 & \cellcolor[rgb]{ .71,  .792,  .902}0.0389 & \cellcolor[rgb]{ .4,  .573,  .792}0.0041 & \cellcolor[rgb]{ .4,  .573,  .792}0.0041 \\
    \end{tabular}%
    
\vspace{0.4cm}
    \begin{tabular}{rrrrrrr}
    \rowcolor[rgb]{ .353,  .541,  .776}    & \cellcolor[rgb]{ .565,  .69,  .851} & \cellcolor[rgb]{ .773,  .835,  .922} & \cellcolor[rgb]{ .988,  .988,  1} & \cellcolor[rgb]{ .984,  .8,  .812} & \cellcolor[rgb]{ .98,  .604,  .612} & \cellcolor[rgb]{ .973,  .412,  .42} \\
    0.01  & 0.05 & 0.22 & 1  & 2.15  & 4.64  & 10 \\
    \end{tabular}%
\end{center}
\caption{Mean von Neumann entropy produced by density matrix models with Fuzz and Phaser on short-form dataset. Colour indicates the relative change in entropy with composition. Note that Phaser caused a greater entropy reduction than Fuzz.}
\label{tab:vne2}
\end{table}

\section{Discussion and Conclusions}
Interpreting metaphor correctly is an essential part of language understanding. Previous work showed that density matrix approaches to modelling language meaning could effectively capture ambiguity. The aim of this paper was to test whether these representations could capture ambiguity in the case of conventional metaphor. We built a new dataset to test our models, and tested a range of density matrix word representations and composition methods, as well as neural sentence encoders. All models found this dataset extremely difficult. However, we do see some insights into how metaphor can be interpreted. Firstly, in almost all compositional methods, we see an increase in performance over a simple verb-only baseline. We also see some increases in performance over simple vector-based models. We find that using the non-metaphorical nouns as operators rather than the metaphorical verbs gives improved performance, indicating that perhaps we need to view the metaphorically used words as being updated by their context. We also see that in the case of Fuzz and Phaser composition, the composition reduces the entropy of representations, as previously seen in \cite{Piedeleu2015, meyer2020}

\paragraph{Further Work}
We have tested our dataset on SBERT and InferSent, however, recently, a wide range of language models have become available, and we plan to test those models on this difficult dataset. The work we have presented in this paper relates only to conventional metaphor, which can be seen fairly straightforwardly as a case of lexical ambiguity. Work is ongoing to extend these methods to novel metaphor, where new meanings must be created on-the-fly. Another line of enquiry is to link the interpretation of density matrices as modelling ambiguity with the interpretation as modelling hyponymy. The categorisation theory of metaphor proposed in \cite{glucksberg1990understanding} argues that in noun-noun metaphors (e.g. `My lawyer is a shark'), an ad-hoc class (e.g. `vicious things') is created that abstracts both the metaphor and its target. We can therefore use the entailment interpretation of density matrices to discover these  ad-hoc classes by finding something like a greatest lower bound for downward entailment -- with the caveat that such bounds are not always unique for density matrices. There is potential to integrate these methods with the DiscoCirc formalism \cite{coecke2021mathematics}, in which a sense of narrative is included, and the level of modelling is at the full text rather than sentence level. Finally, implementing quantum-inspired models using real quantum computers is a burgeoning new line of research \cite{lorenz2023qnlp} and our methods have already been investigated in simulation in \cite{bruhn2021}. Pushing these ideas further will be an important step in this process.

\bibliographystyle{eptcs}
\bibliography{DensityMatrices}

\appendix
\section{Numbers of Sentence Pairs}

\begin{table}
\begin{center}
\begin{tabular}{ c c|c c }

  & & Cosine Similarity & Von Neumann Entropy \\
   \hline
 GloVe & verb-only & 331 \\
& short-form & 269 \\ 
  & long-form & 63 \\ 
 \hline
  Word2Vec & verb-only & 331 \\
  & short-form & 269 \\ 
  & long-form & 63 \\ 
 \hline
   BERT2DM models & verb-only & 300 & 336 \\
  & short-form & 204 & 252 \\ 
  & long-form & 8 & 10 \\ 
 \hline
   Context2DM & verb-only & 310 & 338 \\
  & short-form & 212 & 256 \\ 
  & long-form & 8 & 10 \\ 
 \hline
   Word2DM models & verb-only & 332 & 342\\
  & short-form & 335 & 342 \\ 
  & long-form & 172 & 220 \\ 
\end{tabular}
\end{center}
\caption{Number of sentence pairs used in each test after removing out-of-vocabulary words. SBERT and Infersent covered all words.}
\label{tab:errors}
\end{table}

\end{document}